\documentclass[10pt,twocolumn,letterpaper]{article}

\usepackage[pagenumbers]{cvpr} 

\usepackage{graphicx}
\usepackage{amsmath}
\usepackage{amssymb}
\usepackage{booktabs}
\usepackage{paralist}
\usepackage{bm}

\usepackage[pagebackref,breaklinks,colorlinks]{hyperref}

\usepackage[capitalize]{cleveref}
\crefname{section}{Sec.}{Secs.}
\Crefname{section}{Section}{Sections}
\Crefname{table}{Table}{Tables}
\crefname{table}{Tab.}{Tabs.}

\def\cvprPaperID{*****} 
\def\confName{AI535}
\def\confYear{S2022}

\begin{document}

\title{\ Conditional Image Generation with Pretrained Generative Model
}

\author{Rajesh Shrestha \hspace{20 pt} Bowen Xie \hspace{20pt}  \\
Oregon State University\\
{\tt\small \{xiebo, shresthr\}@oregonstate.edu}
}
\maketitle

\begin{abstract}
   In recent years, diffusion models have gained popularity for their ability to generate higher-quality images in comparison to GAN models. 
   However, like any other large generative models, these models require a huge amount of data, computational resources, and meticulous tuning for successful training. This poses a significant challenge, rendering it infeasible for most individuals. As a result, the research community has devised methods to leverage pre-trained unconditional diffusion models with additional guidance for the purpose of conditional image generative. These methods enable conditional image generations on diverse inputs and, most importantly, circumvent the need for training the diffusion model. In this paper, our objective is to reduce the time-required and computational overhead introduced by the addition of guidance in diffusion models -- while maintaining comparable image quality. We propose a set of methods based on our empirical analysis, demonstrating a reduction in computation time by approximately threefold.
\end{abstract}

\section{Introduction}
Due to a significant potential for a variety of downstream applications, generative models have been very popular, particularly for images and audio. Some of the important generative models include Generative adversarial model\cite{goodfellow2014generative}, VAE \cite{kingma2022autoencoding}, Normalizing flow model\cite{papamakarios2021normalizing}, and autoregressive models \cite{oord2016conditional,lee2022autoregressive}. While GANs have been a go-to generative model for images, their training process can be quite challenging due to the adversarial nature of their loss function. However, following the revelation by \cite{dhariwal2021diffusion} that diffusion models can surpass GANs in terms of generated image quality, they  have emerged as a favored choice of generative model for images. Diffusion models \cite{songdenoising, ho2020denoising} are likelihood-based models and are comparatively easier to train compared to GANs. Their easiness to train and effectiveness in generating high-quality images have led to their wide adoption among the community.

The popularity of AI-generated images and AI-assisted art creation has experienced an enormous surge. Diffusion models have been extensively used in a number of applications such as filling texture based on the edges \cite{zhang2023adding,zhang2023adding}, image generation based on textual and/or location description\cite{nichol2022glide,balaji2023ediffi}, style transfer \cite{balaji2023ediffi}, data augmentation \cite{yuan2023just}, video generation\cite{harvey2022flexible,ho2022video,yang2022diffusion}, image super-resolution \cite{Rombach_2022_CVPR}, image inpainting\cite{lugmayr2022repaint} etc. The success of these applications largely stems from our ability to control their output and tailor it to individual desires. 

A straightforward method for achieving such control is through \emph{conditioning}. While this method is effective, it requires training the diffusion model from very scratch for each specific modality of user input. Consequently, each trained model is tied to a specific type of user input, necessitating the retraining of the entire model with the change in input modality. A more practical approach involves using a method called \emph{guidance} within the diffusion model. In this method, an unconditional pretrained diffusion model acts as a generator which is guided by a guidance function that takes into account both user input and the generated image. The purpose of this guidance function is to approximate the alignment of the generated image with the individual's input.

Given the inherent time-consuming nature of image generation in the diffusion model -- due to the iterative and sequential nature of denoising -- the guidance procedure aggravates this issue. The work in this paper will make use of text-based guidance using CLIP model\cite{radford2021learning} for image generation. Our guidance mechanism is based on \cite{bansal2023universal}. Our approach involves analyzing the significance of various additional steps introduced by the guidance process. The insights gained from this analysis are then leveraged for the purpose of reducing the time required for guided image generation while minimizing image degradation. To assess the quality of the generated image, we will use visual inspection along with \emph{Frechet Inception Distance} (FID) score\cite{heusel2018gans}. 

The following sections first present essential background information on the diffusion model and guidance, followed by an exploration of related works, our methodology, and their corresponding results.


\begin{figure*}[!ht]
  \centering
  \begin{subfigure}{1.0\linewidth}
    \includegraphics[width=\linewidth]{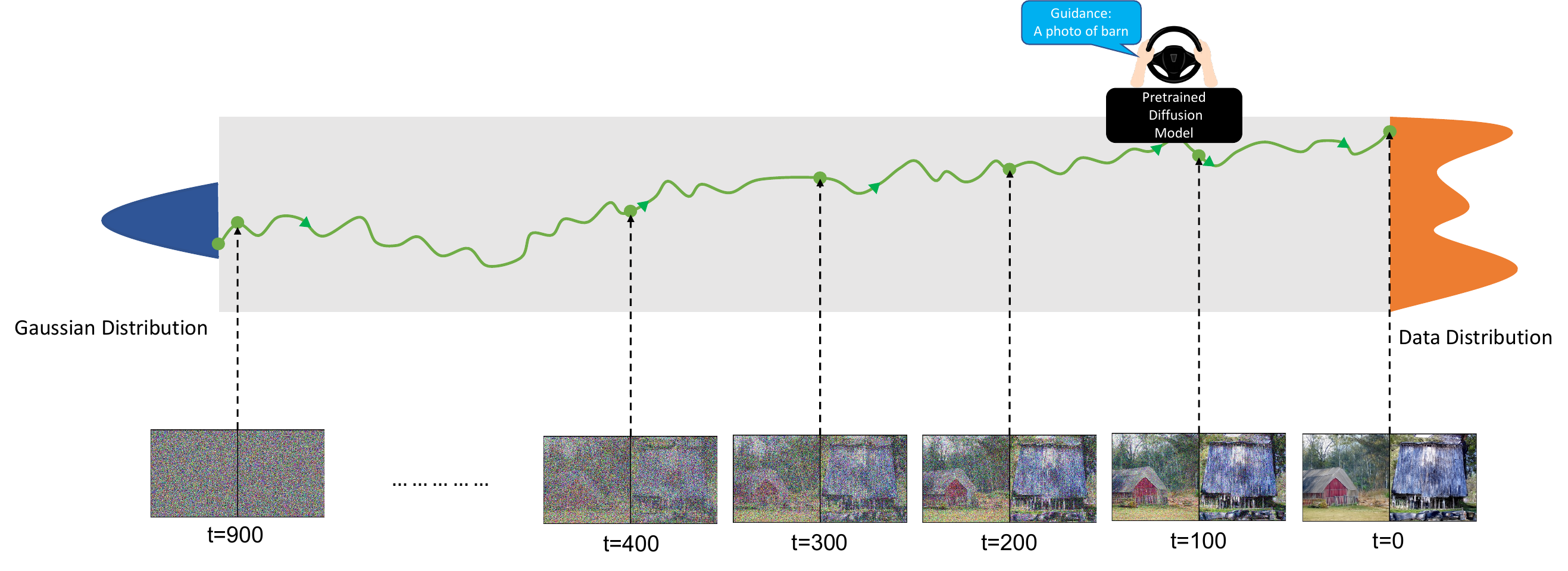}
  \end{subfigure}
    \centering
    \caption{Guidance in Diffusion Model}
    \label{fig:guidance in diffusion model}
\end{figure*}

\section{Background}
In this section, we begin by providing a concise introduction to the basic architecture of diffusion models followed by controlled image generation.
\subsection{Diffusion Model}
Diffusion models are very powerful generative models \cite{ho2020denoising, songdenoising} which were initially proposed for image generation. Here we introduce the unconditioned diffusion model, which can be extended to the conditioned one with guidance.

The diffusion model can be roughly defined as a process involving $T$-step forward process and $T$-step backward process. In each forward step, we add a small Gaussian noise to the previous step data, where the noise added at different steps are independent of each other. We start with a clean image $\bm{z}_0$ and iteratively add noise to it for $T$ steps resulting in $\bm{z}_0,\bm{z}_1,\bm{z}_2,\ldots,\bm{z}_{T-1}$ images.

Precisely speaking, assume we have a clean image $\bm{z}_0$, and given a sequence $\{\alpha_t \}_{t \in \{1 \sim T\}}$ to represent the noise scales at each step, after $t$ steps, we get:
\begin{equation}
    \bm{z}_t = \sqrt{\alpha_t}\bm{z}_0 + (\sqrt{1-\alpha_t})\bm{\epsilon}, \bm{\epsilon} \sim \mathcal{N}(\bm{0}, \bm{I})
\end{equation}

The diffusion model is to use deep neural networks to approximate the denoising process at each backward step. Such deep neural network-based denoising model $\boldsymbol{\epsilon_\theta}$ is trained with the pair of images at a step and the previous step.
\begin{equation}
    \bm{\epsilon}_{\bm{\theta}}(\bm{z}_t, t) \approx \bm{\epsilon} = \frac{\bm{z}_t - \sqrt{\alpha_t}\bm{z}_0}{\sqrt{1 - \alpha_t}}
\end{equation}
Equivalently, we want to approximate $p(\bm{z}_{t-1}|\bm{z}_t)$ using $\boldsymbol{\epsilon_\theta}(\bm{z}_t, t)$.

\subsection{Controlled Image Generation}
Image generation in the diffusion model can be guided either with \emph{conditioning} or \emph{guidance}. In this paper, we study guided image generation which will be introduced in the later section. Here, we will briefly introduce guidance functions used for guided image generation. Consider a differentiable guidance function $h$, and given a particular prompt $c$, we define a loss function $l(.,.)$ which measures the distance between the given prompt $c$ and the generated image $\bm{z}_t$ in some common space. The aim here is to generate an image $\bm{z}_t$ aligned with the prompt $c$ such that $l(c,h(\bm{z}_t)) \approx 0$. 

\section{Related Work}
In previous work, \cite{dhariwal2021diffusion} employed a classifier for image generation conditioned on specific class $c$, but this necessitated training the classifier on noisy images obtained during the reverse diffusion process. Subsequently, the research community proposed classifier-free guidance methods; however, these methods required training class conditional diffusion models \cite{ho2022classifier}. \cite{bansal2023universal} provided a general comprehensive framework for guidance that is compatible with various user input modalities and doesn't require training of any component. This universal framework offers a flexible framework for guiding the generation process without the need for any training. As per our knowledge, past works have put emphasis only on the quality of guided images and ignored the computational overhead placed by the addition of guidance mechanism. 

Here we follow the categorization in \cite{bansal2023universal} dividing prior works into two categories, conditional image generation and guided image generation. The essential difference between these two categories lies in that the first category requires training new diffusion models, and the latter does not need to train new diffusion models which largely reduces the time complexity.

\subsection{Conditional Image Generation}
A very classical work of this kind is \cite{ho2022classifier}. This paper proposed a method called the classifier free guidance method using classifier labels as prompts. We know from the basic diffusion model that a critical step is to approximate $p(\bm{z}_{t-1}|\bm{z}_t)$, and in the classifier free method we try to approximate $p(\bm{z}_{t-1}|\bm{z}_t, y)$ where $y$ is the classifier label we use. This new conditional diffusion model requires training.

\subsection{Guided Image Generation}
Figure \ref{fig:guidance in diffusion model} provides an illustrative depiction of the method employed in this study. We will introduce two key methods for guiding image generation in this section. The first approach, known as classifier guidance, which is initially proposed by \cite{dhariwal2021diffusion}. This method involves the addition of a scaled gradient during the denoising process to enhance the result. However, this method has some drawbacks. Firstly, it is limited to a specific class context, necessitating the retraining of the classifier with the addition of each new class context using noisy images.
To overcome these limitations, the universal guidance method was proposed by \cite{bansal2023universal}that provides an effective solution. This approach serves as a generalization of the classifier guidance method. It allows for the utilization of various form of context without the need for any training. Nonetheless, it has the downside of requiring increased computation and time requirements. This is attributed to the addition of a backward guidance step that involves a separate optimization problem that needs to be solved at each step and a per-step self-recurrence step.

\section{Methodology}
Our work in the paper aims to analyze and improve one of the recent guidance methods for the pretrained diffusion model introduced called Universal Guidance \cite{bansal2023universal}. This method is very flexible and works with a variety of guidance functions and doesn't require any training. However, as there is "no free lunch", this method comes with the price of two computationally intensive steps -- \emph{backward guidance} and \emph{Per-step Self-recurrence}.

The \emph{backward guidance} involves solving an optimization problem  at each step and the \emph{Per-step Self-recurrence} involves the guidance(both forward and backward) to be repeated multiple times in a single step of generation. The following subsections first analyze how this guidance method is dependent on these two computationally expensive substeps. Based on this experiment, we propose our idea for the improvement of the time required by this method. This is done with the attempt to either circumvent and/or mitigate these computational overheads introduced by guidance on top of the normal reverse diffusion process. The following subsections discuss our analysis and improvement methods.

First, let's define some of the notations that will be useful for our method and discussion below. Let $\mathcal{D}$ be the original trained unconditional diffusion model and  $T$ be the total number of diffusion steps involved in $\mathcal{D}$. Let $\bm{z}_t$ be the image at step $t$ which means $\bm{z}_0$ is the clean image and $\bm{z}_{T-1}$ is a random noise. Let $c$ denote the guidance value for the generation. There is a loss function $l(c,h(\bm{z}_t))$ where $h$ is some differentiable function. This loss function measures the alignment of the generated image at step $t$ with the guidance value. This guides the generation process of images.

Let's denote some of the hyperparameters of the Universal guidance method on which our methodology depends. Let $k$ be the number of \emph{Per-step Self-recurrence} steps and $m$ be the number of gradient descent steps for \emph{backward guidance} in each self-recurrence. The guidance procedure using $\mathcal{D}$ would involve $T$ diffusion step. Each such step will have $k$ recurrence steps and each such recurrence step will have 1 forward guidance step, 1 backward guidance step with $m$ gradient steps, and 1 noise addition step. In total, it will have $N_r = T \cdot k$ number of \emph{self-recurrence step}. This leads to $N_r = T \cdot k$ number of forward guidance steps, $N_r \cdot m = T \cdot k \cdot m$ number of backward guidance gradient steps.

\subsection{Dependence on $k$ and $m$}
The number of each operation in this guidance is highly dependent on $k$ and $m$. We expect, the value of $k$ to have a dramatic impact on the running time of this algorithm. The effect of $m$ might not be at the level of $k$, however, its reduction can still bring significant savings on the computation time. 

Reducing $m$ and $k$ will definitely result in less image-generation time, however, might degrade the quality of the image generated based on the fidelity and match with desired guidance. Careful consideration needs to be made while adjusting these values; the aim is to reduce the image generation time as much as possible without sacrificing much on the desired quality of the generated image. We will first study the impact these hyperparameters have on the quality of the image and then, see how the computation time change by adjusting these values. With this experiment, we hope to find a proper value of $k$ and $m$. 

\subsection{Dependence on Guidance at different steps}
Without any guidance, each step during image generation in a diffusion model involves predicting a noise image that could be subtracted from the input image -- to denoise it. This is done with a deep neural network-based denoiser model ($\bm{\epsilon_\theta}$). More steps get added to this normal denoising step to steer the image generation toward the desired one. Inarguably, this adds far more computation in the generation step compared to the unguided one. However, this is necessary if we want to guide the generation.

As with $k$ and $m$, we study the impact of this guidance at different stages of the image generation process. We perform two experiments on this. The first experiment involves changing the guidance at a certain stage of the generation process and analyzing how the final image is impacted by it. Secondly, a kind of ablation study in which guidance is turned off/on after a certain step in the generation. This will provide useful insights into whether the overhead introduced by the guidance can be avoided at certain stages of the diffusion model.

\subsection{Model Based Omission of Backward Guidance and Per-step Self-recurrence} \label{subsection: model based omission}
\begin{figure*}[t]
  \centering
  \begin{subfigure}{0.48\linewidth}
    \includegraphics[width=\linewidth]{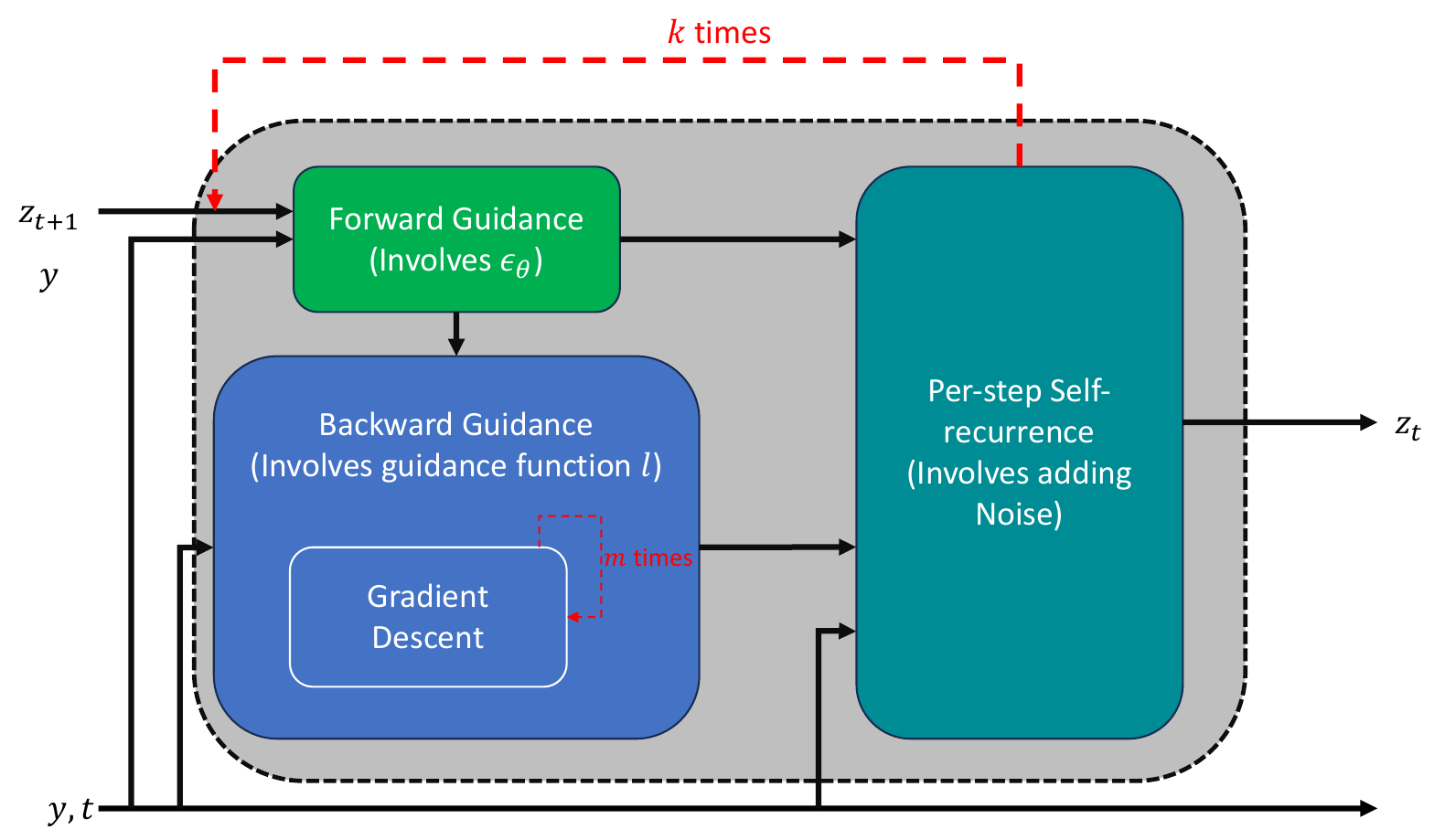}
    \caption{Step in Universal Guidance}
    \label{fig:universal guidance step}
  \end{subfigure}
  \hfill
  \begin{subfigure}{0.48\linewidth}
    \includegraphics[width=\linewidth]{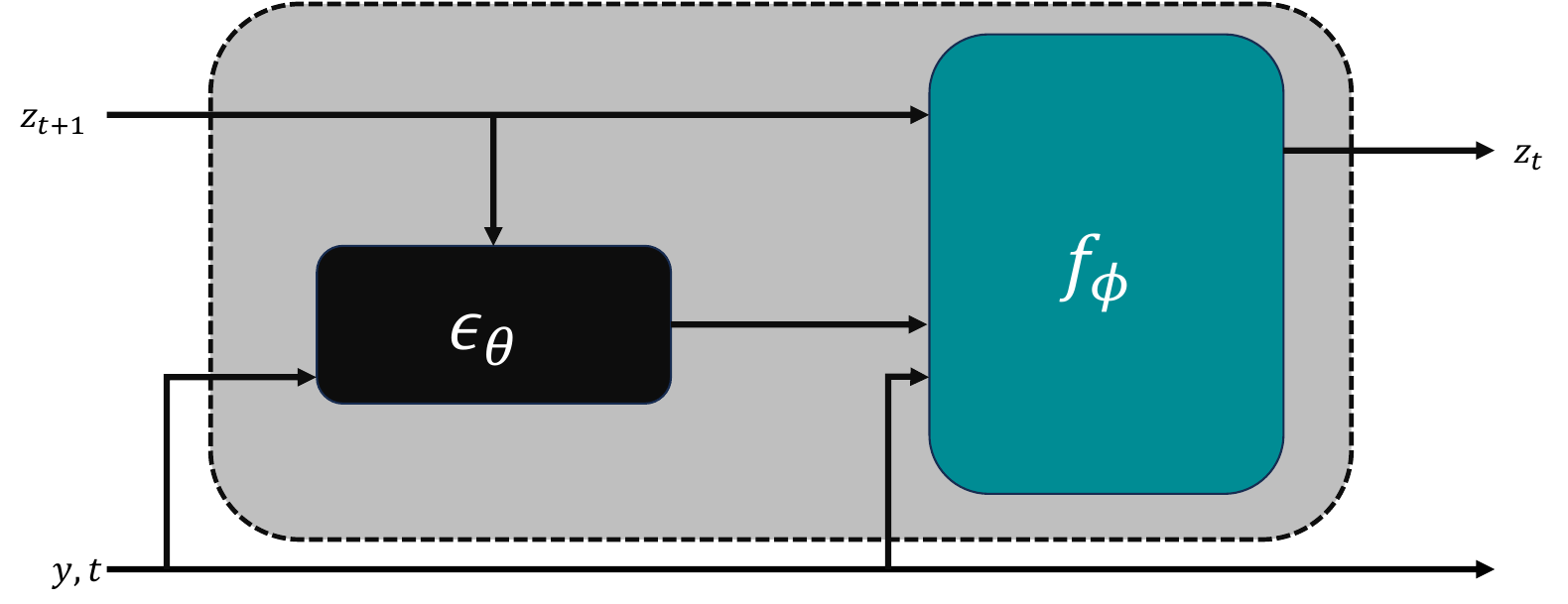}
    \caption{Model based approximation step}
    \label{fig:model based approximation step}
  \end{subfigure}
  \caption{Flow diagram involved in a Step of Diffusion Model}
  \label{fig:flow diagram of step in diffusion model}
\end{figure*}

Our final attempt is to approximate the guidance process with a model-based approach and circumvent the additional steps introduced by the guidance. We train this model $f_\phi$ in a supervised way for which data needs to be collected.

At first, images are generated using a diffusion model with guidance. During this process, at each step, we obtain input image from the previous step $\bm{z}_{t+1}$, the noise estimate in it with $\bm{\epsilon_\theta}(\bm{z}_{t+1})$ and final guided image $\bm{z}_t$. We will use $\bm{z}_{t+1}$, $\bm{\epsilon_\theta}(\bm{z}_{t+1})$, time step $t$, and guidance value $y$ as input for the model $f_\phi$ that tries to predict $\bm{z}_t$. If this can be learned and it generalizes well, then, the guided diffusion will involve two passes to the models -- one for $\bm{\epsilon_\theta}$ and another for $f_\phi$. This way we can avoid the iterative approach currently used in such universal guidance as shown in Figure \ref{fig:flow diagram of step in diffusion model}.

\section{Experiments}
In this section, we provide the details of our experiment setup and present the results obtained from a variety of experiments discussed in our previous section. We show that the computation time required for image generation with guidance in the diffusion model can be significantly reduced without much impact on the quality of generated image and consistency with the guidance.

\subsection{Setup}
\paragraph{Model Architecture:}
The architecture of the diffusion model that we used is similar to the one used in \cite{bansal2023universal}. It is an unconditional diffusion model that uses a U-Net based architecture for the denoiser model $\bm{\epsilon_\theta}$ and DDIM \cite{song2022denoising} based sampling mechanism in each step. We use pretrained model weight\footnote{https://github.com/openai/guided-diffusion} that is trained on $256 \times 256$ colored images of ImageNet dataset \cite{5206848} and provided by OpenAI\cite{dhariwal2021diffusion}.

We used a similar U-Net model for our experiment on model-based approximation discussed in subsection \ref{subsection: model based omission}. This model takes an input image of $256 \times 256 \times 6$ image along with embeddings of time $t$ and text prompt $c$. The input image is obtained by concatenation of $\bm{z}_t$ and $\bm{\epsilon_\theta}(\bm{z}_t)$. An embedding network is learned for time $t$ whereas, clip encoding of $c$ is used in this model.

\paragraph{Clip Guidance: } CLIP \cite{radford2021learning} is one of the popular deep models that connects text to images on an embedding space. It consists of both image and text encoder that maps them to a common embedding space. In this common embedding space, a comparison of a text can be made with an image. 

Our experiments are based on text-guided image generation. For this, we encode our generated image and the guidance text prompt provided using the CLIP encoder and use the negative cosine distance between their embedding as a loss function($l$) for the guidance. From here on, we will refer to such guidance as \emph{clip guidance}.

\paragraph{Computing Resources Used:} The entire experiments were run in the \emph{HPC cluster} provided by \emph{Oregon State University}. Since the aim of our work is to analyze and improve the computation time, we stuck with machines of similar specifications. We used machines with a $15$ core CPU, $200G$ RAM, and A40 GPU for the entirety of this work so the time taken by guided diffusion in different settings is comparable to each other.

\subsection{Results}
\begin{figure*}[!ht]
    \centering
    \begin{subfigure}{0.83\linewidth}
        \includegraphics[width=\linewidth]{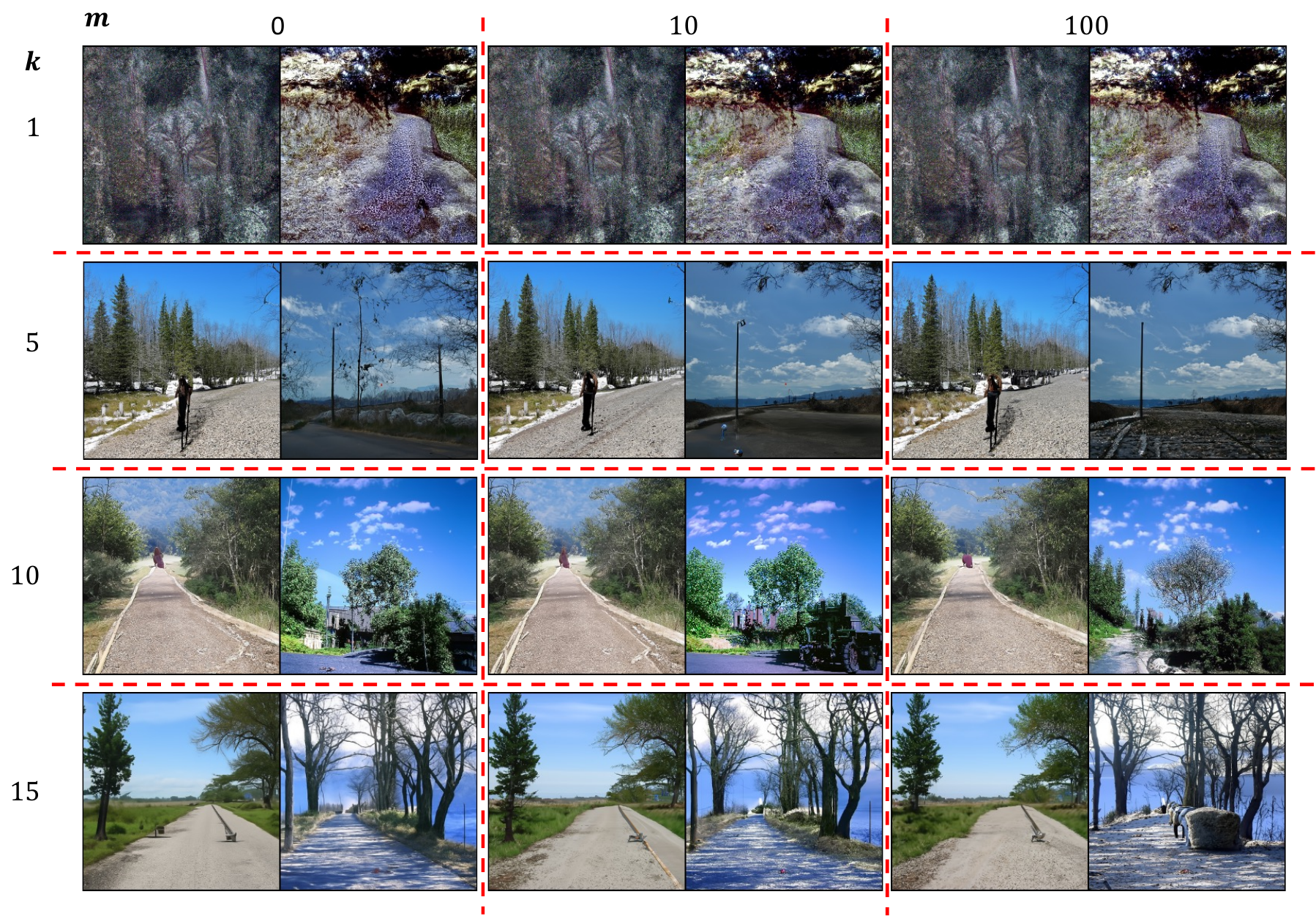}
        \caption{\textbf{Text Prompt: }"A fantasy photo of a lonely road"}
    \end{subfigure}
    \hfill
    \begin{subfigure}{0.83\linewidth}
        \includegraphics[width=\linewidth]{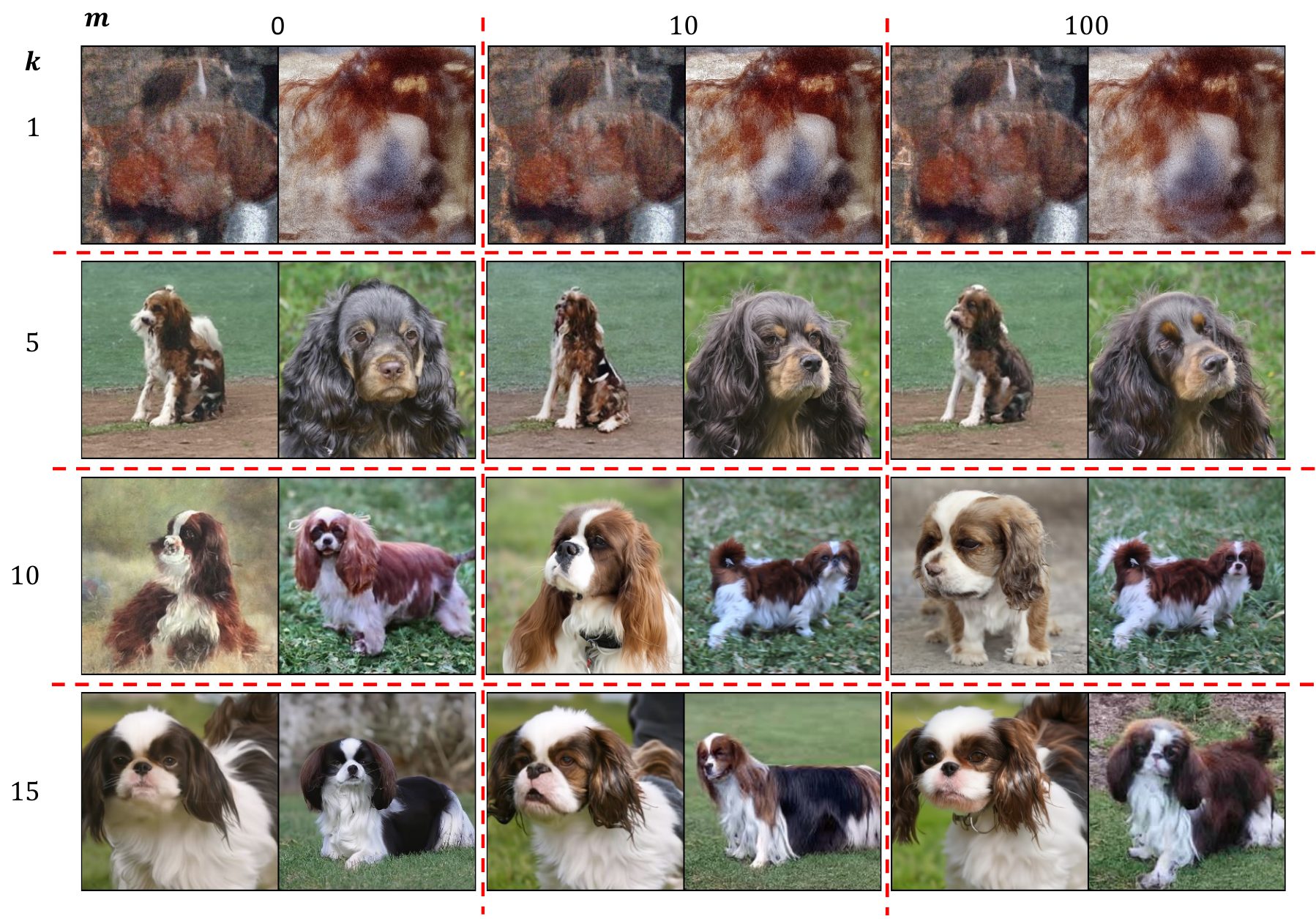}
        \caption{\textbf{Text Prompt: }"A photo of King Charles spaniel"}
    \end{subfigure}
    \centering
    \caption{Effect of $k$ and $m$ on generated image quality}
    \label{fig:effect of k and m on generated image quality}
\end{figure*}

\paragraph{Dependence on $k$ and $m$:}
In this experiment, we compared the images generated with different values of $k=1,5,10,15$ and $m=0,10,15$ for a variety of text prompts. Some of the results are depicted in Figure \ref{fig:effect of k and m on generated image quality}. Our result shows that normal forward and backward guidance in themselves aren't sufficient to generate guided images (corresponds to $k=1$ in the figure). The self-recurrence step makes the generated image more realistic while the forward guidance makes it more consistent with the text. The backward guidance helps to add details to the generated images. The images get more detailed and clearer as the value of $k$ and $m$ increases. However, the improvement in the quality and consistency of generated images isn't drastic. The image generated with $k=5$ seems to be comparable with the ones generated with a higher value of $k$.

\begin{table}[t]
  \centering
  \begin{tabular}{l c c c}
    \toprule
     & $m=0$ & $m=10$ & $m=100$ \\
    \midrule
    $k=1$ & 6:45 & 7:08 & 7:10\\
    $k=5$ & 33:51 & 35:07 & 35:45\\
    $k=10$ & 1:08:25 & 1:13:34 & 1:18:18\\
    $k=15$ & 1:44:26 & 1:51:01 & 2:06:13\\
    \bottomrule
  \end{tabular}
  \caption{Time taken for image generation(HH:MM:SS)}
  \label{table: km time taken}
\end{table}

\begin{figure*}[!ht]
    \centering
    \begin{subfigure}{0.83\linewidth}
        \includegraphics[width=\linewidth]{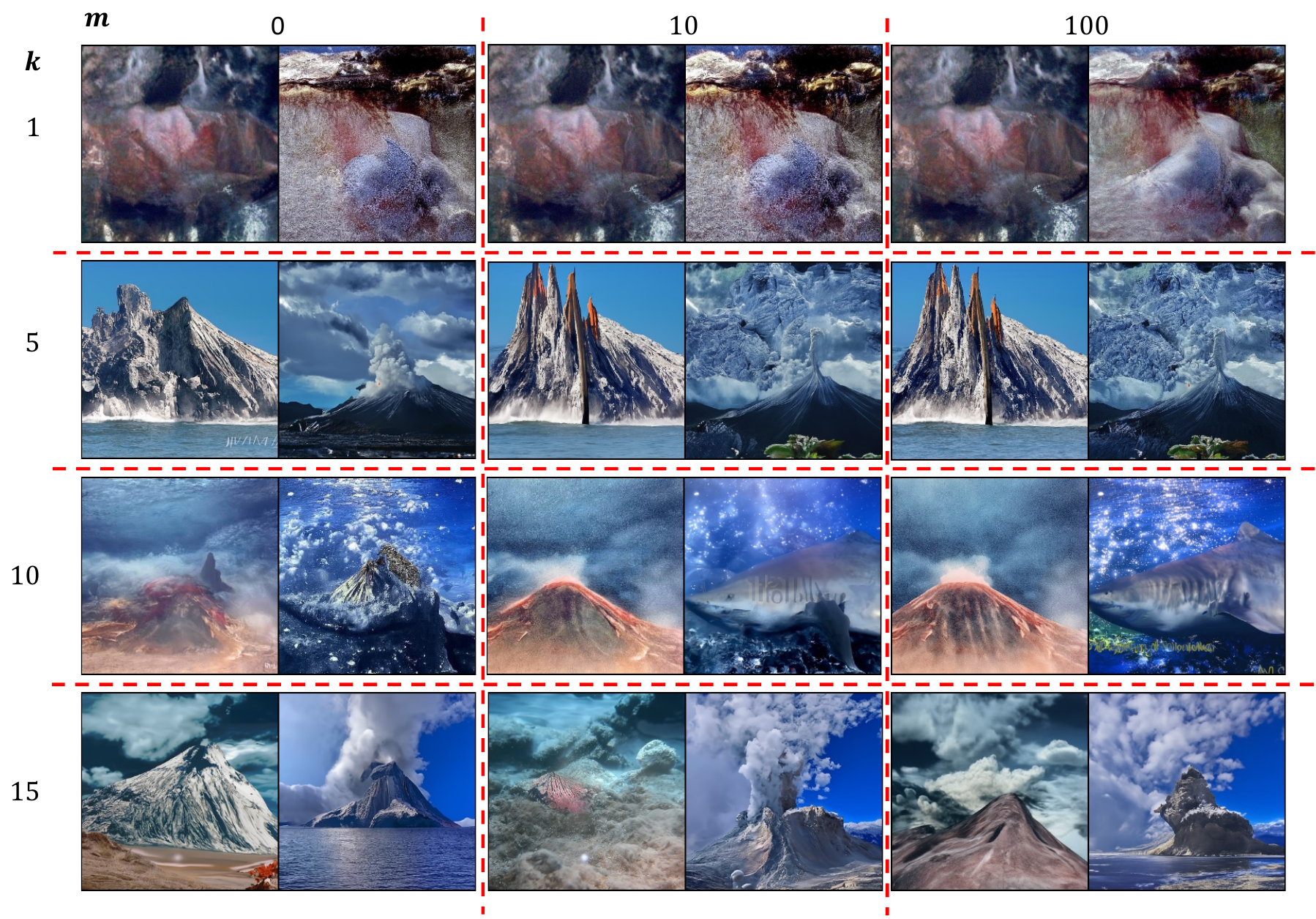}
        \caption{\textbf{Text Prompt: }"A fantasy photo of volcanoes". Artifacts of shark can be observed at $k=10$.}
    \end{subfigure}
    \hfill
    \begin{subfigure}{0.83\linewidth}
        \includegraphics[width=\linewidth]{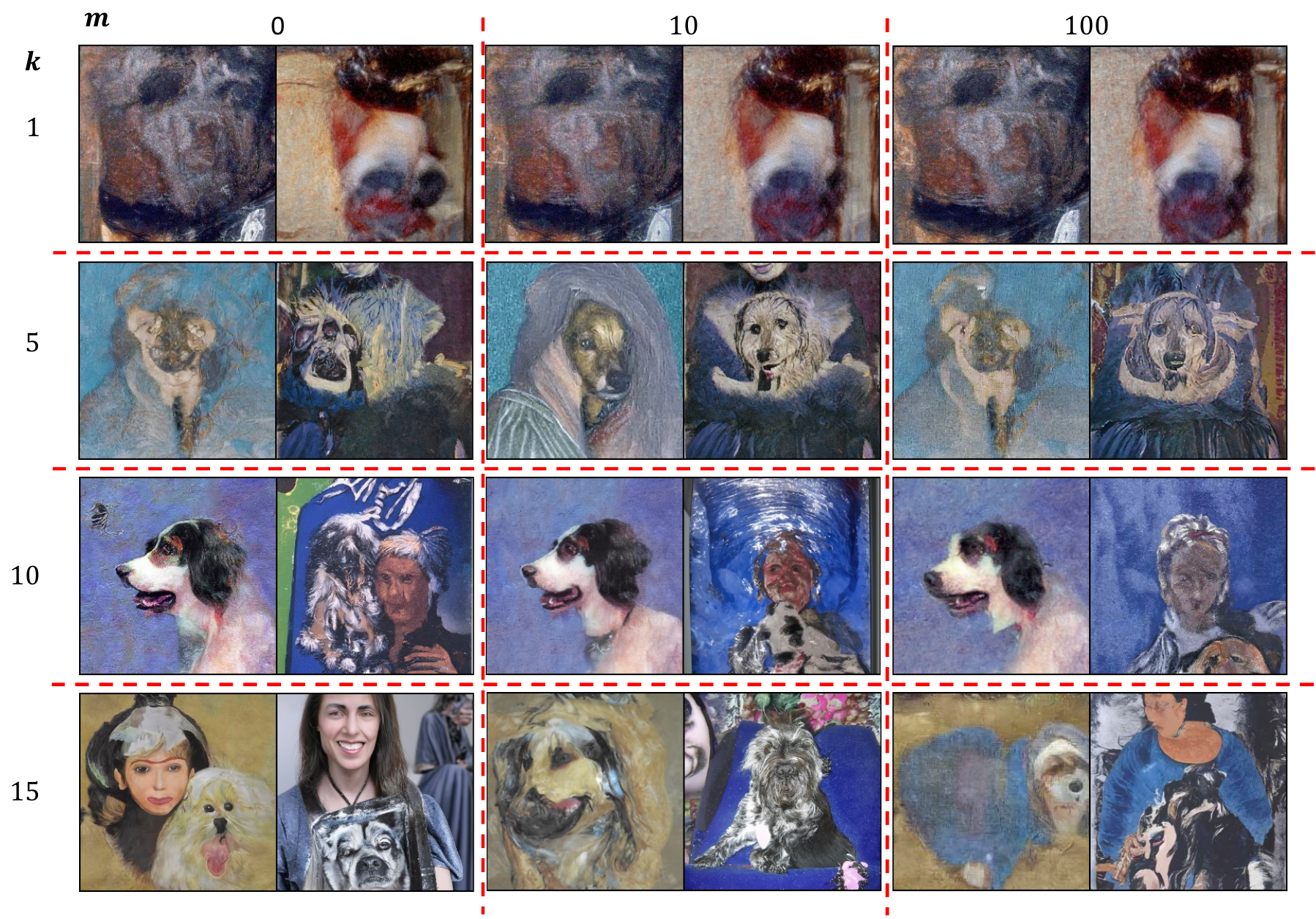}
        \caption{\textbf{Text Prompt: }"An oil painting of a headshot of a woman with a dog". Inconsistent with the text prompt. Real human face and dog generated instead of an oil painting at $k=15$.}
    \end{subfigure}
    \centering
    \caption{Artifacts with high value of $k$}
    \label{fig:artifacts with high value of k}
\end{figure*}

However, the time taken increases rapidly as the number of $k$ increases with a slight increment with the value of $m$ (Table \ref{table: km time taken}). The time taken seems to increase proportionally as the value of self-recurrence $k$ increases. This is expected as all the computation gets increased by a factor of $k$ itself. From the perspective of time required, the setting of $k=5$ and $m=10$ appears to be best without having to compromise much on the image quality and consistency. This claim is further fortified with FID scores of generated images (Table \ref{table: km dependence FID}). The FID score was computed on the images generated at different configurations of $k$ and $m$. \footnote{The precomputed statistics obtained from http://bioinf.jku.at/research/ttur/ was used. The FID implementation of https://github.com/bioinf-jku/TTUR was used in this experiment.} The configuration of $k=5$ seems best in terms of image quality with a low FID score in comparison. This is because sometimes a high value of $k$ leaves weird artifacts in the generated image. It also can compromise its consistency with the text. Such generated images can be seen in Figure \ref{fig:artifacts with high value of k}.

\begin{table}[t]
  \centering
  \begin{tabular}{l c c c}
    \toprule
     & $m=0$ & $m=10$ & $m=100$ \\
    \midrule
    $k=1$ & 338.026 & 341.976 & 365.073\\
    $k=5$ & 285.492 & 285.6265 & 286.722\\
    $k=10$ & 302.717 & 302.114 & 293.892\\
    $k=15$ & 300.601 & 293.3997 & 305.784\\
    \bottomrule
  \end{tabular}
  \caption{FID Score computed with ImageNet using precomputed statistics (lower is better)}
  \label{table: km dependence FID}
\end{table}

\paragraph{Dependence of Generation on Guidance:}
In this, we examine the necessity of guidance at different steps of image generation in the diffusion model. For this, we conducted two analyses.
\begin{enumerate}
    \item \emph{Switching Guidance Value:}
        \begin{figure*}[t!]
            \centering
            \begin{subfigure}{1\linewidth}
                \includegraphics[width=\linewidth]{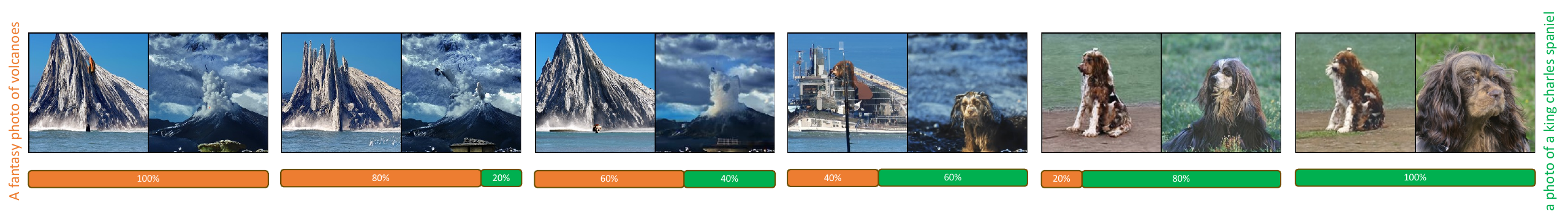}
                \caption{\textbf{Text Prompt: }"A fantasy photo of volcanoes" $\rightarrow$ "a photo of a king charles spaniel"}
            \end{subfigure}
            \hfill
            \begin{subfigure}{1\linewidth}
                \includegraphics[width=\linewidth]{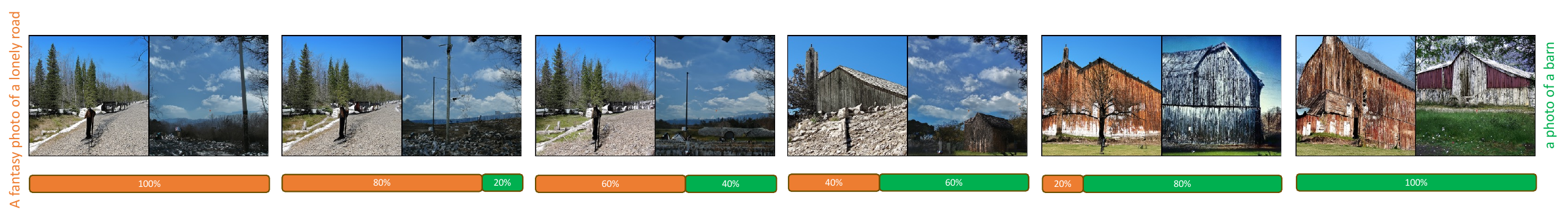}
                \caption{\textbf{Text Prompt: }"A fantasy photo of volcanoes" $\rightarrow$ "a photo of a king charles spaniel"}
            \end{subfigure}
            \centering
            \caption{Impact of changing guidance text at different step on final image}
            \label{fig:impact of changing guidance text at different step on final image}
        \end{figure*}
        We started the image generation process with text $c_1$ and then after a certain $p$ proportion into the reverse diffusion steps, we switched to different text $c_2$ for guidance. We present two samples of results obtained on this experiment in Figure \ref{fig:impact of changing guidance text at different step on final image}. Two things are to be noted in the images. First of all, changing guidance for some of the steps either at the start or end of the process still generates desired images. Secondly, the guidance at the start of the generation is more important compared to the end. Even if we change the text for the last $40\%$ of the steps, an image consistent with $c_1$ is generated. In comparison, in order to generate images consistent with $c_2$, we need to switch the guidance no later than $20\%$ into the generation process. This suggests being able to disable the guidance after $60\%$ into the generation process and follow a normal reverse diffusion step without guidance thereafter.

    \item \emph{Ablation of Guidance}
    \begin{figure*}[t]
        \centering
        \begin{subfigure}{1\linewidth}
            \includegraphics[width=\linewidth]{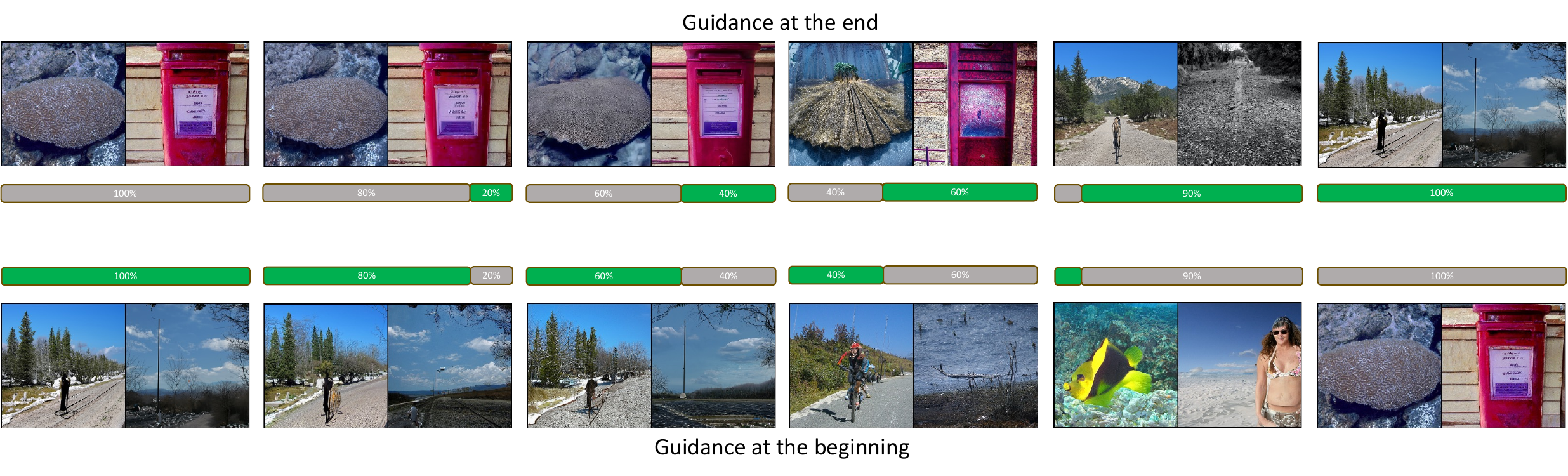}
            \caption{\textbf{Text Prompt: }"A fantasy photo of a lonely road"}
        \end{subfigure}
        \hfill
        \begin{subfigure}{1\linewidth}
            \includegraphics[width=\linewidth]{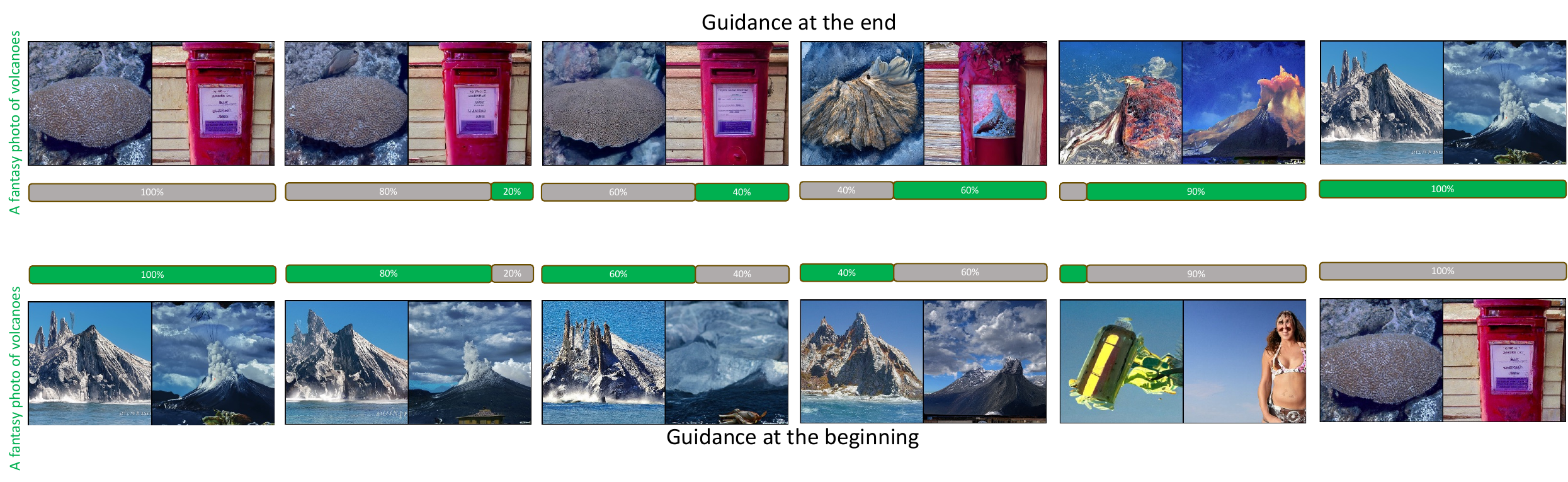}
            \caption{\textbf{Text Prompt: }"A fantasy photo of volcanoes"}
        \end{subfigure}
        \hfill
        \begin{subfigure}{1\linewidth}
            \includegraphics[width=\linewidth]{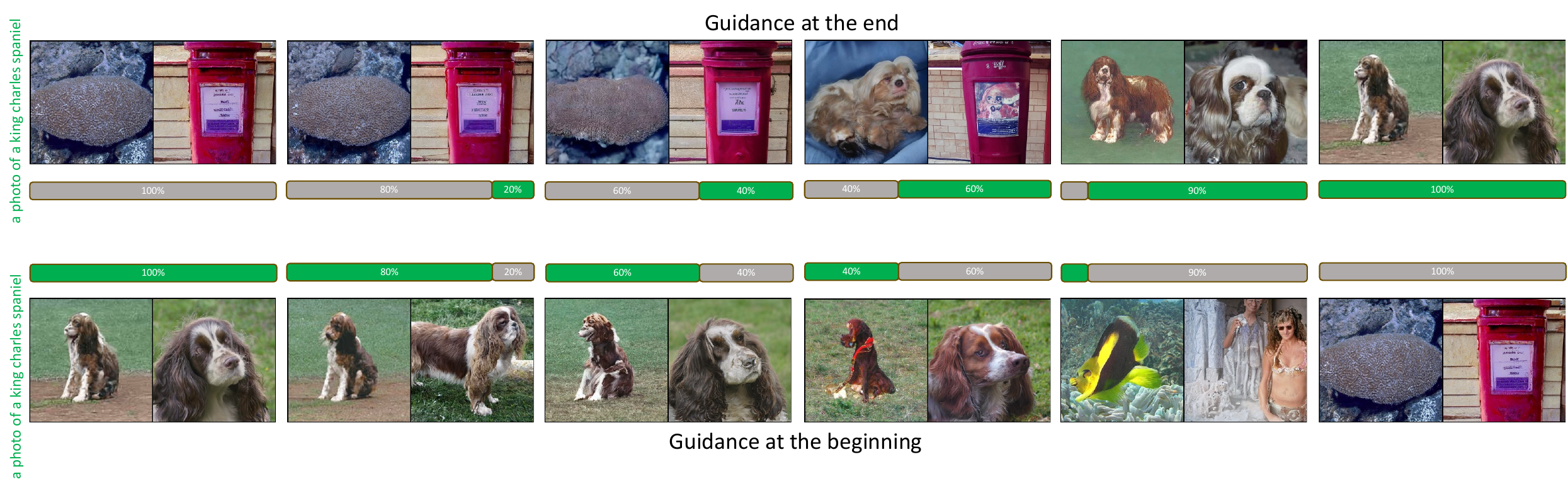}
            \caption{\textbf{Text Prompt: }"a photo of a king charles spaniel"}
        \end{subfigure}
        \centering
        \caption{Guidance Ablation Study. \emph{Top row:} The guidance turned off at the start and activated only after $p\%$ into the generation process. \emph{Bottom row:} Guidance turned on for $p\%$ portion of the generation process at the beginning and turned off after that.}
        \label{fig:guidance ablation study}
    \end{figure*}

    To corroborate the observation from the previous experiment, we performed an additional ablation study. Here, we generated images by either turning the guidance off/on after a certain step into the generation process. The results shown in Figure \ref{fig:guidance ablation study} are consistent with both observations from the previous experiment. The figures clearly depict that initial guidance is important compared to the end. Even with guidance for only the first $60\%$, consistent images are generated. The quality of the image obtained in such a manner is comparable with that of full guidance. As for using the guidance only at the end, we require guidance in about the last $90\%$ to obtain the image of similar quality.

    \begin{table}[t]
      \centering
      \begin{tabular}{c c}
        \toprule
         Guidance Activation Step & time(seconds) \\
        \midrule
        $-0.8$ & $1741.457$\\
        $-0.6$ & $1364.083$\\
        $-0.$4 & $969.646$\\
        $-0.$1 & $413.206$\\
        $1$ & $192.699$\\
        $0.8$ & $581.487$\\
        $0.6$ & $968.18$\\
        $0.4$ & $1371.128$\\
        $0.1$ & $1933.996$\\
        $0$ or $-1$ & $2531.91$\\
        \bottomrule
      \end{tabular}
      \caption{Time taken based on guidance ablation step. The positive value $p$ means the guidance was inactive in the beginning and was activated after $p$ proportion of reverse diffusion steps. In contrast, $-p$ means the guidance was active at the start and turned off after $p$ proportion into reverse diffusion steps.}
      \label{table: guidance ablation time taken}
    \end{table}
    If we compare the time required, using guidance for the first $60\%$ (corresponds to -0.6 in the table) takes about $1364.083$ seconds whereas using guidance the whole time(0 or -1 in the table) takes about $2531.91$ seconds. The computation time is reduced by $\approx 45\%$ without much compromise in the generated image quality.

    \end{enumerate}

\subsection{Model Based Omission of Backward Guidance and Per-step Self-recurrence}
Due to time and resource limitations, we were able to dump about 30K data samples by running the normal guided reverse diffusion on a variety of text prompts. The data was split 6:3:1 for training validation and test data. We trained our UNet based model on this dataset with simple mean square error(MSE) as the loss function and ADAM\cite{kingma2017adam} as the optimizer. Even using GPU with VRAM of $45G$, we were only able to use a batch size of $6$. This is due to the model used being big and the size of each input image being $256 \times 256 \times 6$. 
\begin{figure}[t]
  \centering
  \begin{subfigure}{1.0\linewidth}
    \includegraphics[width=\linewidth]{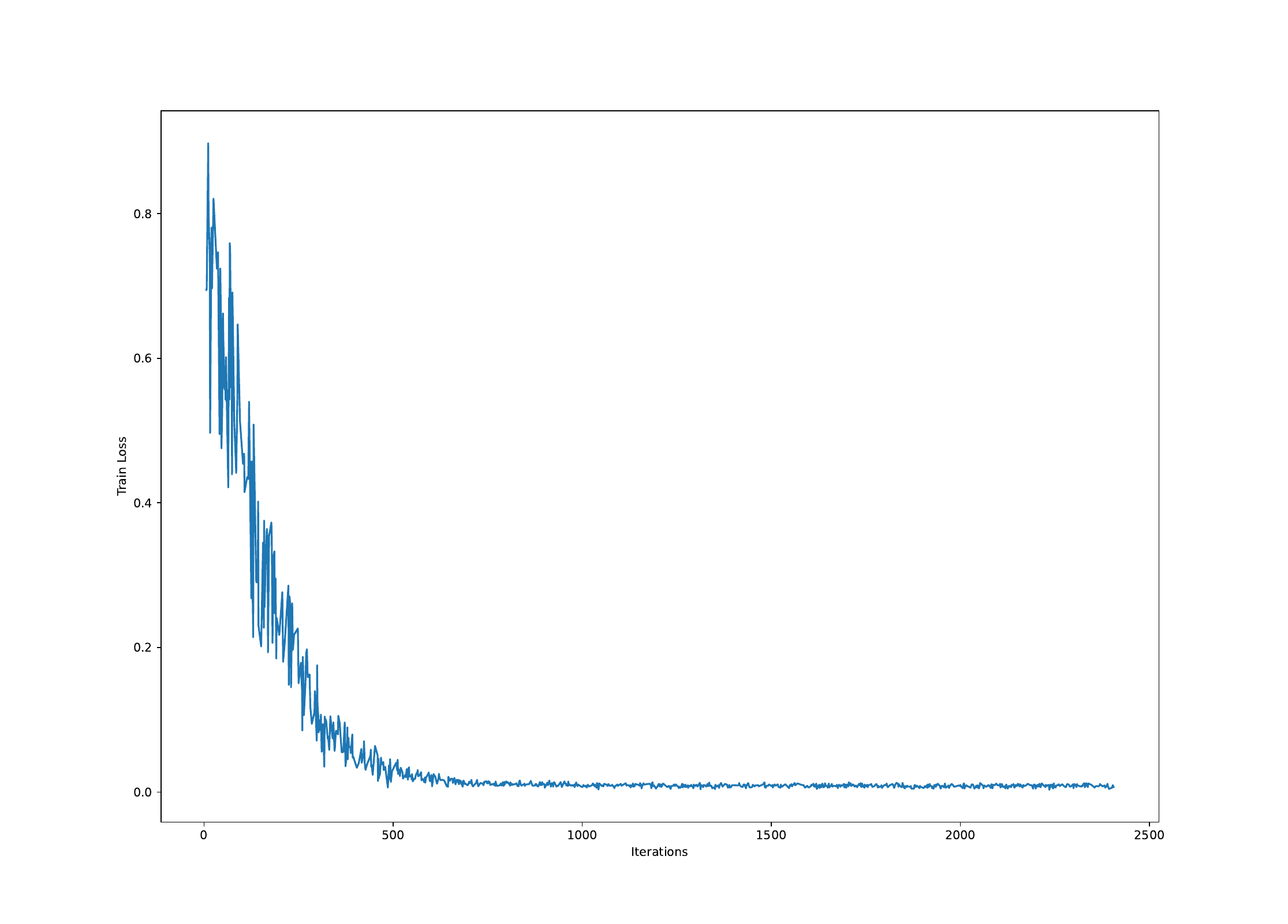}
  \end{subfigure}
    \centering
    \caption{Training Plot of Guidance Step Approximation Model}
    \label{fig:training plot of guidance step approximation Model}
\end{figure}

\begin{figure}[t]
  \centering
  \begin{subfigure}{0.45\linewidth}
    \includegraphics[width=\linewidth]{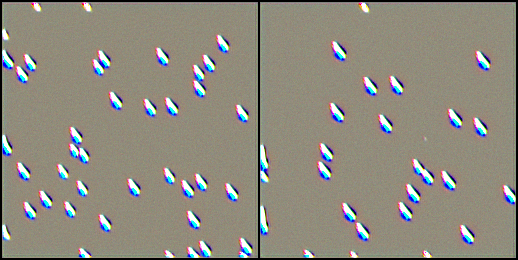}
  \end{subfigure}
    \begin{subfigure}{0.45\linewidth}
    \includegraphics[width=\linewidth]{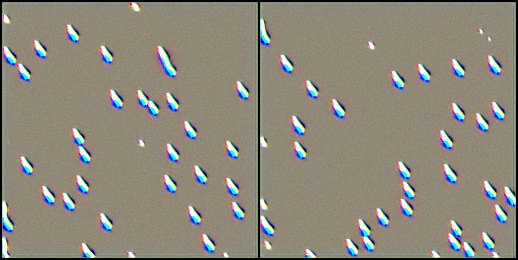}
  \end{subfigure}
    \centering
    \caption{Generated image using $f_\phi$}
    \label{fig:generated image using fphi}
\end{figure}

However, using this model $f_\phi$ during the generation process didn't work. It only produced patches in the generated images even with a variety of text prompts as in Figure \ref{fig:generated image using fphi}. The reason might be a lack of variety of data on a diverse set of captions. Maybe the model didn't perform well on some of the steps which messed up the whole generation process. Further analysis and experimentation are required for this.

\section{Conclusion}
The diffusion model coupled with the guidance mechanism offers a promising and effective approach for controlled image generation. This offers an interesting research area with numerous practical applications. Nevertheless, the inclusion of guidance introduces additional computational overhead to an already time-consuming iterative generation process in the diffusion model. In this work, we conducted a comprehensive analysis of the components of a popular guidance mechanism and proposed methods to mitigate the overhead posed by the addition of guidance. Through our experiments, we demonstrated that our proposed methods are effective at reducing time and result in significant time reduction, up to a factor of 3, without significantly compromising image quality. While our model-based approximation didn't yield a satisfactory result, we remain hopeful that with further engineering refinements and potentially a large dataset, this can be made to work. These aspects are reserved for exploration in future work.

Although our experiments showed promising ways to reduce the time requirement, it is important to  note that our test wasn't performed in a large set of generated images. Therefore, more rigorous testing is needed to validate our findings. However, we are very optimistic with our methods as we obtained consistent results across images generated with various texts. Additionally, it is crucial to extend on diverse forms of guidance and their potential impact on image generation beyond the text guidance using the CLIP model.

{\small
\bibliographystyle{ieee_fullname}
\bibliography{egbib}
}

\end{document}